\documentclass{article}

\usepackage{microtype}
\usepackage{graphicx}
\usepackage{subfigure}
\usepackage{booktabs} 

\usepackage{hyperref}

\usepackage{algorithm}
\usepackage{algorithmicx}
\usepackage{algpseudocode}
\usepackage{multirow}

\usepackage{enumitem}

\usepackage[preprint]{neurips_2025}
\usepackage[utf8]{inputenc} 
\usepackage[T1]{fontenc}    
\usepackage{hyperref}       
\usepackage{url}            
\usepackage{booktabs}       
\usepackage{amsfonts}       
\usepackage{nicefrac}       
\usepackage{microtype}      
\usepackage{xcolor}         
\usepackage{amsmath}
\usepackage{amssymb}
\usepackage{mathtools}
\usepackage{amsthm}
\usepackage{wrapfig}

\definecolor{commentcolor}{HTML}{93a1a1}
\usepackage[capitalize,noabbrev]{cleveref}

\theoremstyle{plain}

\theoremstyle{definition}

\theoremstyle{remark}

\newcommand{\src}{\mathrm{src}}
\newcommand{\tgt}{\mathrm{tgt}}

\usepackage[textsize=tiny]{todonotes}

\title{Discrete Noise Inversion for Next-scale Autoregressive Text-based Image Editing
}

\author{Quan Dao$^{1\dagger*}$\qquad\quad Xiaoxiao He$^{1\dagger}$\qquad\quad Ligong Han$^{2*}$ \\ \textbf{Ngan Hoai Nguyen}$^{3}$\qquad\quad \textbf{Amin Heyrani Nobari}$^{4}$\qquad\quad \textbf{Faez Ahmed}$^4$\qquad\quad \\ \textbf{Han Zhang}$^{5}$\qquad\quad \textbf{Viet Anh Nguyen}$^{6}$\qquad\quad \textbf{Dimitris Metaxas}$^{1}$ \\
$^1$Rutgers University\hspace{1cm}$^2$Red Hat AI Innovation\hspace{1cm}$^3$Independent Reseacher\\$^4$MIT\hspace{1cm}$^5$ReveAI\hspace{1cm}$^6$CUHK\\
}

\begin{document}

\maketitle

\begin{abstract}
Visual autoregressive models (VAR) have recently emerged as a promising class of generative models, achieving performance comparable to diffusion models in text-to-image generation tasks. While conditional generation has been widely explored, the ability to perform prompt-guided image editing without additional training is equally critical, as it supports numerous practical real-world applications. This paper investigates the text-to-image editing capabilities of VAR by introducing Visual AutoRegressive Inverse Noise (VARIN), the first noise inversion-based editing technique designed explicitly for VAR models. VARIN leverages a novel pseudo-inverse function for argmax sampling, named Location-aware Argmax Inversion (LAI), to generate inverse Gumbel noises. These inverse noises enable precise reconstruction of the source image and facilitate targeted, controllable edits aligned with textual prompts. Extensive experiments demonstrate that VARIN effectively modifies source images according to specified prompts while significantly preserving the original background and structural details, thus validating its efficacy as a practical editing approach.
\end{abstract}

\section{Introduction}
\label{sec:intro}

In the era of generative AI, diffusion models \citep{ho2020denoising, song2020score} have emerged as dominant methods in image synthesis, outperforming GANs \citep{goodfellow2014generative} in both \textit{un}conditional and conditional generation tasks \citep{dhariwal2021diffusion}, notably in text-to-image generation \citep{rombach2022high}. This success has enabled diverse practical applications, including personalization \citep{ruiz2023dreambooth, van2023anti, kumari2022customdiffusion}, 3D generation \citep{poole2022dreamfusion, wang2023prolificdreamer}, and prompt-based image editing \citep{brack2024ledits, huberman2024edit, mokady2023null, cyclediffusion, hertz2022prompt, he2024dice}, underscoring their increasing popularity and utility.
\footnote{${\dagger}$: Equally Contribution, ${*}$: Corresponding Author}
In contrast, autoregressive models, traditionally dominant in natural language processing, have only recently gained traction in visual synthesis. Recent models such as LlamaGen \citep{sun2024autoregressive} and MagVit-v2 \citep{yu2023language} have optimized image tokenization and transformer architectures, achieving performance competitive with diffusion models. Furthermore, MARS \citep{he2024mars} integrates mixtures of experts to train large-scale text-to-image autoregressive models, whereas MAR \citep{li2024autoregressive} removes vector quantization by leveraging diffusion-inspired training in continuous space. Despite achieving diffusion-level quality, these methods still incur significant inference costs due to dependence on output size, limiting scalability for high-resolution generation. Addressing this issue, \cite{tian2024visual} proposed next-scale prediction to substantially reduce inference time without sacrificing performance, highlighting potential future advantages of autoregressive approaches. Additionally, \cite{tang2024hart} introduced the Hybrid VAR Transformer (HART) diffusion framework, combining visual autoregressive models (VAR) with lightweight diffusion refinement, achieving comparable results to pure diffusion methods but with reduced inference time. This model raises further research opportunities in downstream tasks, including personalization, prompt-based image editing, and text-to-3D synthesis within VAR-based frameworks.

\begin{figure}[t]
    \centering
    \includegraphics[width=1\linewidth]{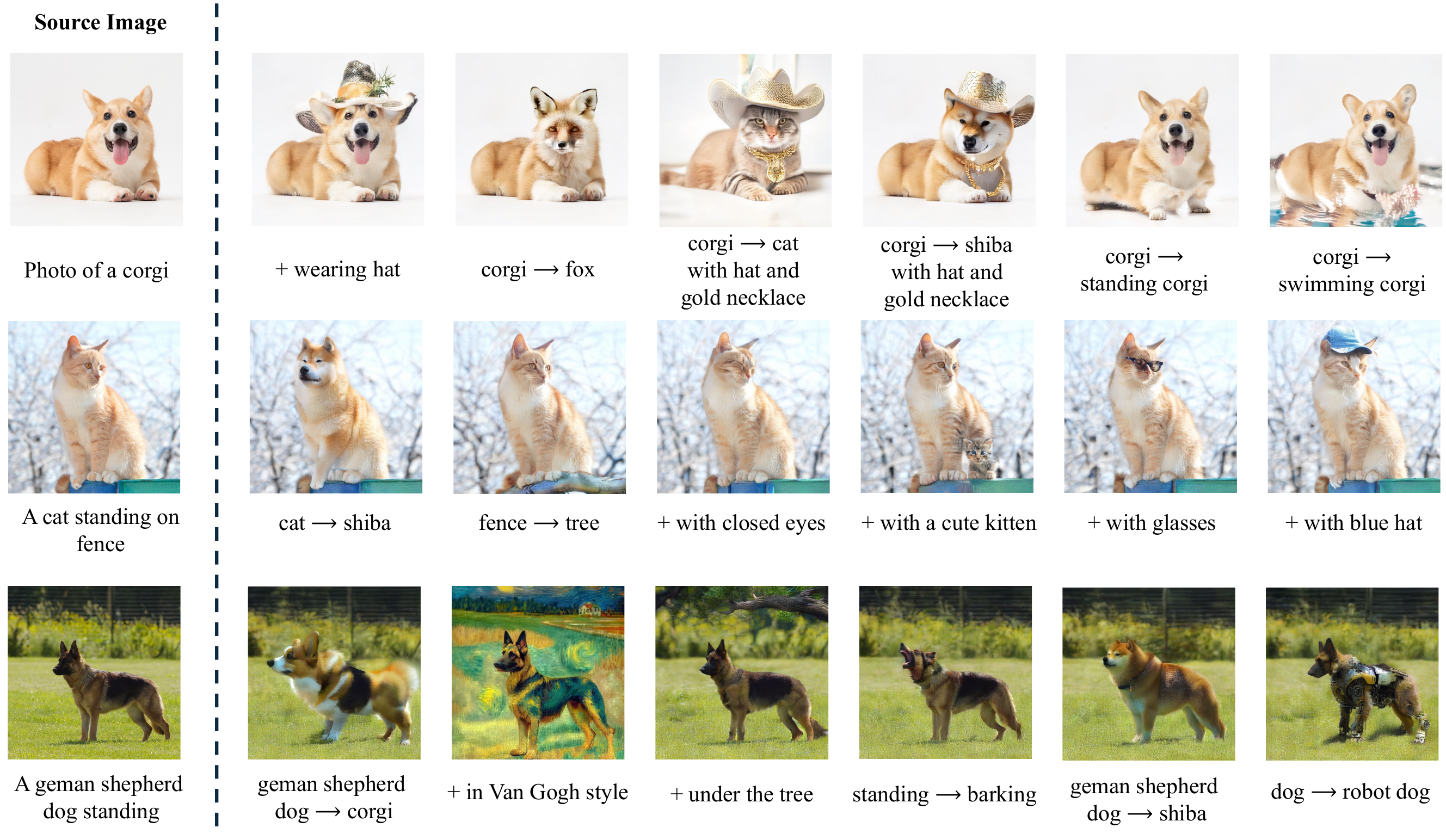}
    \vspace{-4mm}
    \caption{Qualitative performance of VARIN given diverse prompts.}
    \label{fig:diverse}
    \vspace{-0.7cm}
\end{figure}

In this paper, we focus specifically on prompt-guided text-to-image editing within visual autoregressive models (VAR). Given a source image and a text prompt describing desired edits, the task is to modify the image according to the prompt while preserving unrelated details from the original. We observe that VAR generates the overall structure and layout at initial scales, progressively refining finer details at higher scales. As illustrated in \cref{fig:hart}, editing primarily becomes necessary at middle scales, around levels 6 or 7. A straightforward baseline is \textbf{Regeneration}, which preserves tokens at initial scales and regenerates remaining scales. However, this approach inadequately retains non-target image details.

To address these limitations, we propose \textbf{Visual Autoregressive Inverse Noise (VARIN)}, the first training-free editing method designed explicitly for VAR. Inspired conceptually by noise inversion techniques in diffusion models, VARIN identifies editable noise sets capable of perfectly reconstructing the source image, enabling precise editing. While continuous autoregressive models (e.g., Gaussian-based) allow straightforward inversion through inverse transformation flows \citep{kingma2016improved}, discrete VAR models like \citep{tian2024visual, tang2024hart} complicate inversion due to reliance on the non-invertible argmax sampling (Gumbel-max trick). The simplest pseudo-inverse, the one-hot argmax inversion \citep{he2024dice}, produces uncontrollable noise sets, limiting precise editing capabilities.
Therefore, we propose \textbf{Location-aware Argmax Inversion (LAI)}, a novel pseudo-inverse function enhancing controllability and alignment with target prompts. LAI extracts inverse noises to reconstruct source images precisely and allows adjustable bias towards preserving source details, significantly outperforming basic regeneration. Our method demonstrates editing quality on par with more complex, optimization-based test-time tuning approaches such as Null-Text Inversion, without requiring additional retraining or intricate cross-attention manipulations. These advanced methods remain complementary and could be combined in future work for further improvements.
We summarize our contributions as follows:
\begin{itemize}[nosep,leftmargin=*]
\item We introduce first editing technique for VAR named VARIN. Our VARIN is based on noise inversion technique to obtain set of editable noises and control these noises for editing image. 
    \item We propose Location-aware Argmax Inversion (LAI) as pseudo-inverse of argmax inversion. This allows us to extract inverse noises that perfectly reconstruct the source image. Furthermore, we can control the bias information of source image in extracted inverse noises, leading to better results.
    \item We validate the effectiveness of VARIN through experiments on text-to-image editing tasks, demonstrating their ability to align with target prompts while preserving source image details.
\end{itemize} 

\vspace{-0.1cm}

\section{Related Work}\label{sec:related}
\noindent \textbf{Visual autoregressive models.}
Autoregressive models have significantly shaped NLP, particularly through their next-token prediction paradigm~\citep{vaswani2017attention, achiam2023gpt, touvron2023llama, chowdhery2023palm, workshop2022bloom, bai2023qwen, team2023gemini}. This same paradigm has also proven effective in visual generation: models such as VQVAE~\citep{van2017neural, razavi2019generating}, VQGAN~\citep{esser2021taming, lee2022autoregressive}, DALL-E~\citep{ramesh2021zero}, LlamaGen~\citep{sun2024autoregressive}, and MARS~\citep{he2024mars} tokenize images for next-token prediction. While these token-based methods can match diffusion models in image quality, their inference speed often suffers because the number of tokens grows with image resolution.

Recent work shifts toward ``next-scale prediction,'' where models predict multiple size scales in parallel via residual quantization~\citep{lee2022autoregressive}. VAR~\citep{tian2024visual} and HART~\citep{tang2024hart} exemplify this approach by generating high-quality images in only a few steps. Furthermore, VAR is closely related to ImageBART~\citep{esser2021imagebart}, which applies transformers to each denoising step in multinomial diffusion~\citep{hoogeboom2021argmax}; one can interpret VAR similarly, but using a ``blurring-to-deblurring'' viewpoint.

\noindent \textbf{Diffusion editing.}
Diffusion-based text-to-image editing has gained popularity for its flexibility and controllability~\citep{meng2021sdedit, huberman2024edit, nguyen2024flexedit, huang2024diffusion}. Many works rely on large-scale pretraining~\citep{zhang2023adding, brooks2023instructpix2pix, fu2023guiding, sheynin2024emu, zhang2024hive} or fine-tuning approaches~\citep{han2023svdiff, zhang2023sine,dong2023prompt, shi2024dragdiffusion}, where either model weights or embeddings are optimized at test time to achieve editing. Null-text Inversion~\citep{mokady2023null} further refines editing control by adjusting ``null'' embeddings to align the reconstruction path with the source image.

\noindent \textbf{Diffusion inversion.}
A key challenge in diffusion-based editing is extracting an internal representation from the source image so that edits can be made without losing fidelity. Continuous models often invert via deterministic or stochastic processes~\citep{chen2018neural, song2021denoising, lipman2022flow, liu2022flow, wu2022unifying, huberman2024edit}, while recent research addresses discrete diffusion~\citep{he2024dice} and masked generative models~\citep{chang2022maskgit}. These methods confirm that ``diffusion inversion'' can guide image editing by reconstructing the original data from a learned representation. Inspired by these, our work focuses on inverting a visual autoregressive model to enable text-driven editing while preserving source image content.

\section{Preliminaries}\label{sec:preli}
In this section, we firstly review about the visual autoregressive model in \cref{sec:preli:background}. Motivating from method SDEdit \citep{meng2021sdedit} for diffusion, latter \cref{sec:preli:baseline} introduces simple editing method for VAR called \textbf{Regeneration}, where we simply initialize the VAR generative process by beginning token maps $r_1, r_2, \dots, r_t$ extracted from source image using VAR encoder and generate the rest of token maps $r_{t+1}, r_{t+2}, \dots, r_K$ following target text prompt $c_{\tgt}$. Although this technique produces a fairly good final image that has the same structure as the source image and follows the target text prompt $c_{\tgt}$, they lose background details as the final token maps are completely generated without any constraint from the source image. 

\subsection{Visual Autoregressive Model} \label{sec:preli:background}

\begin{figure}[t]
    \centering
    \includegraphics[width=1.0\linewidth]{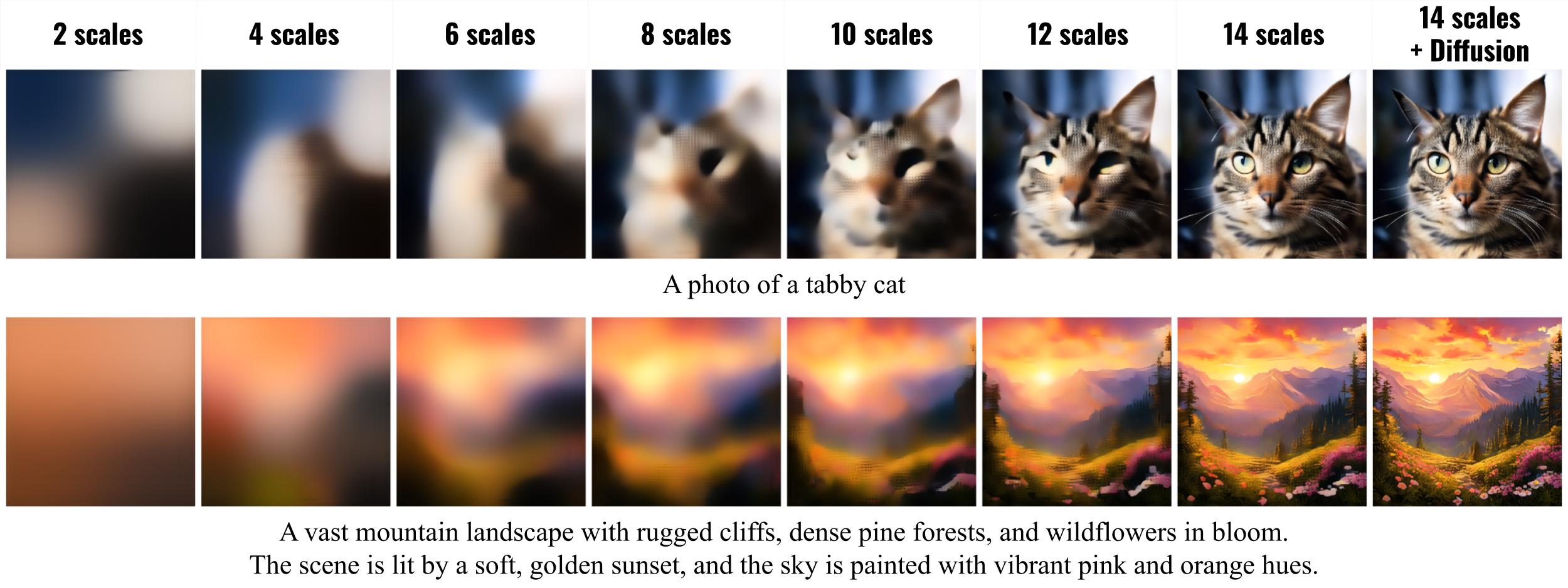}
    \vspace{-4mm}
    \caption{Visualizations of each scale of the generation process of HART~\citep{tang2024hart}. The features of a cat (top) and the landscape (bottom) are only distinguishable above 6 scales.}
    \label{fig:hart}
    \vspace{-0.2cm}
\end{figure}

VAR has a Variational AutoEncoder (VAE) based on VQ-VAE with $[C] = \{1, \ldots,C\}$ of vocal size $C$. In training process, given an image $I \in \mathbb{R}^{3\times H \times W}$, the VAR-VAE encoder $\mathcal{E}_\text{VAR}$ will output the image into $K$ token maps, $(r_1, r_2, \dots, r_K) = \mathcal{E}_\text{VAR}(I)$, where each token maps $r_k$ has different resolution $h_k \times w_k$ and these resolution increases with the scale $k$. The VAR-AutoRegressive model $\theta$ can now be considered as the next scale predictor, which models the following likelihood:
\begin{equation}
    p_\theta(r_1, r_2, \dots, r_K) = \prod_{k=1}^{K}p_\theta(r_k|r_1, r_2, \dots, r_{k-1}), \label{eq:var}
\end{equation}
where $r_K \in [C]^{h_w \times w_k}$ is the token map at scale $k$, and the sequence $(r_1, r_2, \dots, r_{k-1})$ is the prefix of $r_k$. In inference time, we use VAR-AutoRegressive to predict the sequence $r_1, r_2, \dots, r_K$ iteratively. We then pass the sequence to the VAR-VAE decoder $\mathcal{D}_\text{VAR}$ to construct the generative RGB images $I_{\mathrm{gen}} = \mathcal{D}_\text{VAR}(r_1, r_2, \dots, r_K)$.

Unlike traditional autoregressive methods, which flatten the image following a predefined scanning order then perform the next token prediction, the next scale prediction of VAR strictly follows the autoregressive's behaviour. In VAR, each token map $r_k$ depends only on the previous token maps $r_{< k}$. As a consequence, VAR costs shorter inference time than diffusion and traditional next-token prediction autoregressive models since they only need to run model for only few scale prediction ($K \approx 14$) instead of diffusion and vanilla autoregressive with thousands of model iterations.

\subsection{Text-based Image Editing \& Baseline Regeneration} \label{sec:preli:baseline}
\noindent \textbf{Text-based Image Editing:}
We consider the text-based editing problem for the text-to-image VAR model. Given a source image $I_{\src} \in \mathbb{R}^{3 \times H \times W}$ and a target prompt $c_{\tgt}$, we want to edit the source image $I_{\src}$ to comply with the prompt $c_{\tgt}$. In our proposed VARIN in \cref{sec:method}, we use inversion algorithm for editing. Therefore, we need to define a corresponding source text $c_{\src}$ which algins with source image for noise extraction. The inverse noises are then used for editing process with target prompt $c_\tgt$ as shown in \cref{sec:method:editing}. The image editing is evaluated based on two main criteria, which could be conflicting: the edited image should be \textbf{aligned with the target prompt} while still \textbf{preserve the unedited part from the source image}.

\noindent \textbf{Baseline Regeneration:}
Using the VAR encoder, we could obtain the following token maps $(r_1, r_2, \dots, r_K) = \mathcal{E}_\text{VAR}(I_{\src})$. As illustrated in \cref{fig:hart}, while the token map of the beginning scales builds up the structure and layout of images, the later token map adds more detailed information. 
For editing with diffusion model, SDEdit \citep{meng2021sdedit} starts denoising generative process from middle noise level with some guidance to perform editing, while keeping the overall structure of the source image. Motivating from SDEdit, we could pick a scale index $s$ between $1$ and $K$, then we fix the token maps for the beginning scales from $r_1, \dots, r_s$, and start to generate $r_{s+1}, \dots, r_K$ conditioning on the target prompt $c_{\tgt}$. We name this intuitive editing method \textbf{Regeneration}, and outline it in \cref{alg:regen}.


By keeping the beginning scale token map of the source image, the editing output image $I_{\tgt}$ could keep the overall structure and layout of the source image. However, it may fail miserably to preserve the fine-grained details in background that should not be edited, examples of failures are shown in \cref{fig:main_qualitative}. To search for a more fine-grained editing control, we approach the noise inversion technique and investigate how to control the inverse noise for editing in next section.
\vspace{-0.2cm}
\section{Inverse Autoregressive Transformation and Editable Inverse Noise}
\label{sec:method}
\begin{figure}[t]
    \centering
    \includegraphics[width=\linewidth]{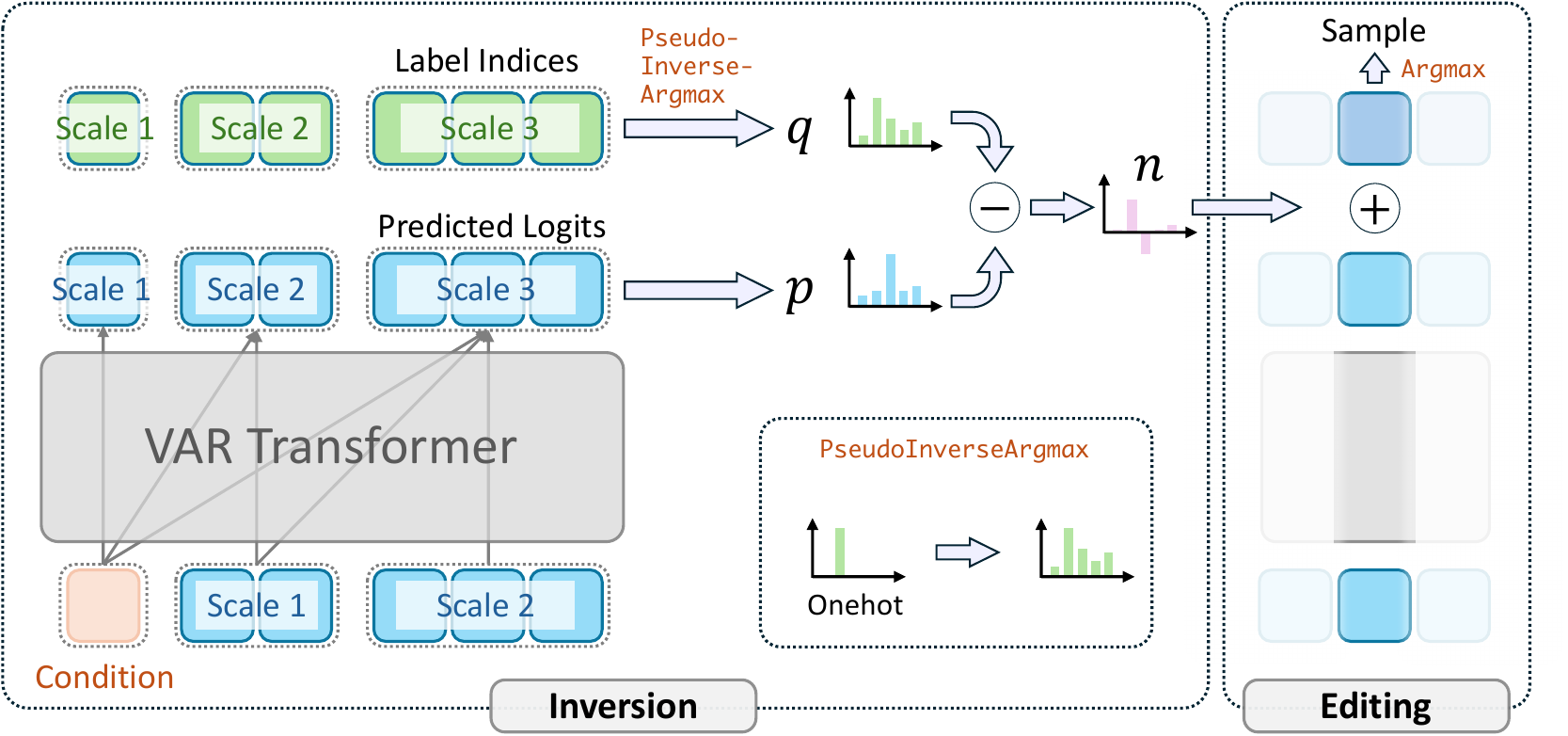}
    \vspace{-0.5cm}
    \caption{VARIN pipeline: the prefix token $c_{src} + r_{<k}$ is fed to transformer to get log probability $p_k$. We then use pseudo inverse-argmax to find the inverse noise $n_k$ from ground truth label $r_k$ and logit $p_k$. These noise set $n_1, n_2, \dots, n_K$ is later used for editing control.}
    \label{fig:method}
    \vspace{-0.5cm}
\end{figure}

We introduce our technique \textbf{Visual AutoRegressive Inverse Noise (VARIN)} that could preserve better the unedited part of the source image. Towards this goal, we discuss the process of noise inversion for an autoregressive model in \cref{sec:method:inverse}. For discrete space, there is no closed-form solution for inversion due to argmax operator, we therefore propose pseudo-inverse function for the argmax operator to obtain inverse noises from source image in \cref{sec:method:pseudo}. Finally, we propose the inverse noise editing algorithm for text-based image editing. The overal pipeline is in \cref{fig:method}.

\subsection{Discrete Inverse Autoregressive Transformations} \label{sec:method:inverse}

Given a continuous Gaussian autoregressive model $p_\theta(x_t|x_{< t})$ and a sequence of tokens $\{x_1, x_2, \dots, x_T\}$, we could apply an inverse transformation in parallel to find the sequence of inverse noises that perfectly reconstruct the sequence~\citep{kingma2016improved}. To see this, note that under the Gaussian assumption of $p_\theta(x_t| x_{<t})$, we have
$x_t = \mu_t + \sigma_t \odot \epsilon_t$, where $\mu_t$ and $\sigma_t > 0$ are the conditional mean and standard deviation of $x_t$ given the history $x_{<t}$. We can obtain $\epsilon_t$ by $\epsilon_t =\frac{x_t - \mu_t}{\sigma_t}$.
Computing the inverse noise under the Gaussian case could be implemented in parallel due to the independence of $\epsilon_t$ and $\epsilon_{t'}$ over different time stamps $t \neq t'$. The above inversion only holds when the noise admits a continuous density. Unfortunately, most of autoregressive generative models for vision including VAR are based on discrete tokens and model the sequence of token with multinominal distribution. Consequentially, the inversion could no longer be straightforwardly integrated into the VAR pipeline.

We now propose a pseudo-inversion for VAR models that extends the continuous noise inversion to the case of \textit{discrete} tokens. The VAR models use a multinomial distribution, where samples are drawn using the Gumbel-max trick. A Gumbel distribution with mean 0 and scale $1$ has a continuous density $\exp\big(-z + \exp(-z)\big)$ for any value $z$. 
We consider the discrete inverse autoregressive transformation problem using the Gumbel-max trick as follows:
\begin{equation} \label{sec:eq:inverse}
    \begin{aligned}
    &p_t \leftarrow p_\theta( \cdot |r_{< t}) \\ 
    &\text{Find Gumbel noise $n_t$ s.t. } \text{argmax}{(p_t + n_t) = r_t}, 
    \end{aligned}
\end{equation}
with (unnormalized) log probability $p_t \in \mathbb{R}^{l \times C}$, $n_t$ is the Gumbel noise from Gumbel-max trick, and $r_t \in \mathbb{R}^{l}$ is the ground truth label with $l$ is number of tokens. As shown in \cref{sec:eq:inverse}, there is no perfect way to obtain the reverse Gumbel noise from a label and predicted probability, as the Gumbel-max trick involves \textbf{an argmax operator}, which is apparently \textbf{non-invertible}. Therefore, in the next section, we propose two pseudoinverse functions for the argmax operator in \cref{sec:method:pseudo}.
\subsection{Pseudo-inverse Argmax} \label{sec:method:pseudo}
Since $(p_t + n_t)$ represents the Gumbel-perturbed logits, we need to choose the unormalized log probability $q_t$ such that $q_t = p_t + n_t$ and $\arg\max(q_t)=r_t$. Notably, to ensure the editing ability, we better need the $n_t$ to follow the below properties:
\begin{itemize}[nosep,leftmargin=*]
    \item \textbf{Prop 1:} $n_t$ needs to be likely sampled from standard Gumbel noise since in line 6 of \cref{alg:edit_noise}, we interpolate $n_t$ with standard Gumbel noise for preserving randomness of generative process.
    \item \textbf{Prop 2:} $n_t$ needs to preserve some bias information from source image. This allows the editing algorithm to preserve the unedited part from source image.
\end{itemize}

Now, we discuss about how to choose pseudoinverse function for argmax operator. The easiest way is setting $q_t = \text{LogOnehot}(r_t)$, and we call this \textbf{Onehot Argmax Inversion (OAI)}. With OAI, we can find the list of inverse noise for perfect reconstruction. However, when using inverse noise OAI for editing, it usually fails to control the editing process. The reason is that the $q_t$ is $0$ at the label index and significantly negative at other indices. This results in $n_t = q_t - p_t$ not likely being sampled from a standard Gumbel noise distribution violating \textbf{Prop 1} and $n_t$ will be highly biased by source image since $q_t$ is extremely biased by label of token. Therefore, in editing, it is hard to control the editing process with OAI. Furthermore, we notice that the OAI process relies only on ground-truth labels and omits the information of predicted logits $p_t$ from models.

\begin{figure}[t]
\centering
\begin{minipage}[t]{0.52\textwidth}
\begin{algorithm}[H]
    \caption{Location-aware Argmax Inversion Function (LAI)}
    \label{algo:LAI}
    \begin{algorithmic}[1]
        \item[\textbf{Input:} tokens $r\in {[}C{]}^l$, log probability $p\in \mathbb{R}^{l \times C}$,]
        \item[\hspace{1cm} and information preserving term $\tau$]
        \State $r_{\text{mask}} \leftarrow \text{Onehot}(r, \text{num\_classes}=C)$
        \Comment{\textcolor{commentcolor}{$\in \mathbb{R}^{l\times C}$}}
        \State $l_{\text{max}} \leftarrow \text{Sum}(r_{\text{mask}} \odot p, \text{dim}=-1)$
        \Comment{\textcolor{commentcolor}{$\in \mathbb{R}^{l \times 1}$}}
        \State $q_{\max} \sim \text{Gumbel}( \mu=l_{\text{max}}, \beta=1)$
        \Comment{\textcolor{commentcolor}{$\in \mathbb{R}^{l \times 1}$}}
        \State $q \sim \text{GumbelTrunc}(\mu=p, \beta=1, \text{trunc}=q_{\text{max}}-\tau)$ 
        \Comment{\textcolor{commentcolor}{$\in \mathbb{R}^{l\times C}$. See~\cref{alg:gumbel_trunc}}}
        \State $q \leftarrow q \odot (1-r_{\text{mask}}) + q_{\text{max}} \odot r_{\text{mask}}$ 
        \Comment{\textcolor{commentcolor}{$\in \mathbb{R}^{l\times C}$}}
        \item[\textbf{Output:} $q$]
    \end{algorithmic}
\end{algorithm}
\end{minipage}%
\hfill
\begin{minipage}[t]{0.47\textwidth}
\begin{algorithm}[H]
    \caption{Visual Autoregressive Inverse Noise (VARIN)}
    \label{alg:ar_inverse}
    \begin{algorithmic}[1]
        \item[\textbf{Input:} Source image $I_\src$, text prompt $c_{\tgt}$]
        \item[\textbf{Parameters:} $\tau$]
        \item[\textbf{Parallel Autoregressive Inversion:}] \label{algo:iar}
        \State $(r_1, r_2, \dots, r_K) \leftarrow \mathcal{E}_\text{VAR}(I_{\src})$
        \For {$t$ from $1$ \text{to} $K$} 
        \Comment{\textcolor{commentcolor}{logits}}
        \State $q_t \leftarrow \text{LAI}(r_t, p_t)$ \label{algo:eq:pseudo}
        \Comment{\textcolor{commentcolor}{Pseudo-inverse argmax~\cref{algo:LAI}}}
        \State $n_t \leftarrow q_t - p_t$
        \Comment{\textcolor{commentcolor}{Gumbel noise inversion}}
        \EndFor
        \item[\textbf{Return:} $(n_1, n_2, \dots, n_K)$]
    \end{algorithmic}
\end{algorithm}
\end{minipage}
\vspace{-20pt}
\end{figure}

To remedy the above issues, we propose a novel argmax inversion function called \textbf{Location-aware Argmax Inversion (LAI)}. Since $q_t = p_t + n_t$ and $n_t$ is sampled from standard Gumbel noise, $q_t$ should be close to $p_t$. LAI exploits this property, and it takes $p_t$ as the location of Gumbel-max sampling. This makes $q_t$ closer to $p_t$, and $n_t$ is more likely sampled from standard Gumbel distribution which is then satisfied \textbf{Prop 1}. For the remaining discussion, we will omit subscript $t$ to avoid clutter notation. To ensure the argmax condition, for each token $i$ from $1$ to $l$, the ground truth label is $r[i]$, we sample $q[i]$ so that $\text{argmax}(q[i]) = r[i]$. Firstly, we sample the value for $q[i]_{r[i]}$ using the Gumbel-max trick with the predicted location $p[i]_{r[i]}$. Later, we sample other position $q[i]_{\neq r[i]}$ by truncated Gumbel-max trick with corresponding predicted location and truncated value $q[i]_{r[i]} - \tau$. Please noted that $\tau$ is very important parameter to control the unedited part preservation and we shall discuss details of $\tau$ below.

\noindent \textbf{Hyperparameter $\tau$ in LAI: } When $\tau = 0$, the $n_t$ from LAI satisfies \textbf{Prop 1} but it still fails to edit while preserving the background. 
We could explain this effect: For simplicity, we consider $q \in \mathbb{R^{C}}$ is a vector. When $\lambda = 0$, the $q$ sampled from \cref{alg:gumbel_trunc} (Gumbel Trunc) is highly uncertain cause there exist some indices $j$ such that $q_j \approx q_i$, where $i = \text{argmax}_k q_k$. Therefore, during editing, 
$
q^{\text{edit}} = p + (1-\lambda) \cdot g + \lambda \cdot n = p + n + (1-\lambda) \cdot (g-n) = q + (1-\lambda) \cdot (g-n) = q + \epsilon.
$
Since $q$ is highly sensitive, even a small $\epsilon$ can cause a change in the maximum index of $q^{\text{edit}}$, such that $\text{argmax}_k q^{\text{edit}}_k = j$. As a result, preserving the unedited part becomes exceedingly difficult. Therefore, by setting $\tau$, we could let $q_t$ retains bias from source image which provides useful information to $n_t$ and control the sensitivity of  \cref{alg:edit_noise} better. This is also satisfied \textbf{Prop 2}.

\begin{wrapfigure}{r}{0.55\textwidth}
  \vspace{-23pt}
  \begin{minipage}{0.55\textwidth}
\begin{algorithm}[H]
    \caption{Editing by VARIN}
    \label{alg:edit_noise}
    \begin{algorithmic}[1]
        \item[\textbf{Input:} ]
        \State $(r_1, r_2, \dots, r_K) \leftarrow \mathcal{E}_\text{VAR}(I_{\src})$
        \State $n_1, n_2, \dots, n_K \leftarrow \text{VARIN}((r_1, r_2, \dots, r_K), c_{\src})$
        \For {$t$ from $s$ \text{to} $K$} \Comment{\textcolor{commentcolor}{$s$ is staring scale}}
        \State $p_t\leftarrow p_\theta(\cdot|r_{<t}, c_{\tgt})$ 
        \Comment{\textcolor{commentcolor}{$p_t$ is log probability}}
        \State $g \sim \text{Gumbel}(0, I)$
        \State $q_t = p_t + (1-\lambda)\cdot g + \lambda n_t $
        \State $\tilde r_t = \text{argmax}(q_t)$
        \EndFor
        \State $I_{\tgt} \leftarrow \mathcal{D}_\text{VAR}(r_1, \ldots, r_{s-1}, \tilde r_s, \ldots, \tilde r_K)$
        \item[\textbf{Output: } $I_{\tgt}$]
    \end{algorithmic}
\end{algorithm}
  \end{minipage}
  \vspace{-40pt}
\end{wrapfigure}

The efficient implementation is shown in \cref{algo:LAI}. LAI always guarantees perfect reconstruction, the same as OAI, but $q_t$ is closer to $p_t$, and $n_t$ is more like standard Gumbel, which satisfied both above noise properties. We use LAI in \cref{algo:eq:pseudo} of \cref{alg:ar_inverse}.

\subsection{Editing with VARIN} \label{sec:method:editing}

For editing using inversion, similar to Regeneration Editing, we first obtain the token maps for each scale using the VAR encoder. We then use \cref{alg:ar_inverse} to collect the list of inverse noise $n_1, n_2, \dots, n_K$. For each $t$, we get the predicted log probability $p_t$ from autoregressive model $p_\theta$. Instead of sampling using the Gumbel-max trick given $p_t$ as in \cref{alg:regen}, which uses new Gumbel noise $g$, we interpolate new noise $g$ with inverse noise $n_t$ by interpolation coefficients $\lambda$. The $\lambda$ is hyperparameter for tuning the editing process.

\section{Experiments}\label{sec:exp}
In this section, we first provide details of the evaluation dataset for text-based image editing task in \cref{sec:exp:dataset}. \cref{sec:exp:recon} presents the reconstruction performance of our proposed method, VARIN + HART. Later, in \cref{sec:exp:editing}, we conduct editing experiments to demonstrate that VARIN can effectively do text-based image editing. Due to limited space, ablation is put in appendix.

\begin{table*}[t]
    \centering
    \vspace{-0.3cm}
        \caption{We evaluate our method VARIN against discrete generative model such as DICE and EditAR. We can see that our method outperforms DICE in terms of editing part with better CLIP Similarity but slightly underperforms in terms of background preservation due to the T2I reconstruction ability. With EditAR, we are better in terms of background reconstruction. Compared to recent continuous diffusion, we also achieve on par performance.}
    \label{tab:quantitative}
    \resizebox{1.0\linewidth}{!}{
    \begin{tabular}{lllccccccc}
    \toprule
    &\multicolumn{2}{c}{\textbf{Method}} &\multicolumn{1}{c}{\textbf{Structure}} &\multicolumn{4}{c}{\textbf{Background Preservation}} &\multicolumn{2}{c}{\textbf{CLIP Similarity}}\\
    
    \cmidrule(r){2-3} \cmidrule(lr){4-4} \cmidrule(l){5-8} \cmidrule(l){8-10}
    
    &\textbf{Name} &\textbf{T2I} &\textbf{Distance$_{\times 10^3} \downarrow$}  &\textbf{PSNR $\uparrow$}     &\textbf{LPIPS$_{^{\times 10^3}}$ $\downarrow$}  &\textbf{MSE$_{^{\times 10^4}}$ $\downarrow$}     &\textbf{SSIM$_{^{\times 10^2}}$ $\uparrow$} &\textbf{Whole $\uparrow$} &\textbf{Edited $\uparrow$}  \\
    \midrule
    \multirow{10}{*}{\rotatebox[origin=c]{90}{{\small \textbf{Continuous}}}}
    &Prompt-to-Prompt    &SD1.4  &69.43	&17.87	&208.80	&219.88	&71.14  &25.01	&22.44\\
    &Negative Prompt     &SD1.4  &16.17	&26.21	&69.01	&39.73	&83.40  &24.61	&21.87\\
    &PnP Inversion       &SD1.4  &\textbf{11.65}	&27.22	&54.55	&\textbf{32.86}	&84.76  &25.02	&22.10\\
    &Null-text Inversion &SD1.4  &13.44	&27.03	&60.67	&35.86	&84.11  &24.75	&21.86\\
    &Pix2pix-zero        &SD1.4 &61.68	&20.44	&172.22	&144.12	&74.67	&22.80	&20.54\\
    &MasaCtrl            &SD1.4 &28.38	&22.17	&106.62	&86.97	&79.67	&23.96	&21.16\\
    &InfEdit             &SD1.4  &14.22  &\textbf{27.52}  &\textbf{47.98}  &34.17  
    &\textbf{85.05}  &24.89  &22.03\\
    &InstructPix2Pix     &SD1.5   &107.43	&16.69	&271.33	&392.22	&68.39	&23.49	&22.20\\
    &MGIE                &SD1.5    &67.41	&21.20	&142.25	&295.11	&77.52	&24.28	&21.79\\
    &DDPM-Inversion                &SD1.4     &22.12	&22.66	&67.66	&53.33	&78.95	&\textbf{26.22}	&\textbf{23.02}\\
    \midrule
    \multirow{4}{*}{\rotatebox[origin=c]{90}{{\small \textbf{Discrete}}}}
    &DICE                &Paella &\textbf{11.34} &\textbf{27.29} &\textbf{52.90} &\underline{43.76} &\textbf{89.79} &23.79 &21.23 \\
    &EditAR              &LlamaGen  &39.43	&21.32	&117.15	&130.27	&75.13	&\underline{24.87}	&\textbf{21.87}\\
    &Regeneration        &HART  &25.56&20.45	&106.50	&95.90	&73.31    &24.65	&21.13\\
    &VARIN               &HART  &\underline{11.46}&\underline{26.54}	&\underline{54.04}	&\textbf{38.33}	&\underline{85.39} &\textbf{25.05}	&\underline{21.49}\\
    \bottomrule
    \end{tabular}
    }
\end{table*}

\label{sec:exp:dataset} 
\noindent\textbf{Dataset: }To conduct the experiment, we use the Prompt-based Image Editing Benchmark (PIE-Bench) \citep{ju2023direct} which is a dataset created to evaluate text-to-image (T2I) editing methods. It includes 700 images in 9 different editing scenarios, making it a useful evaluation protocol for testing how well these methods handle text-guided image edits. The dataset contains detailed annotations and variety of tasks for us to thoroughly test and fairly compare our method with other approaches. 

\subsection{Inversion Reconstruction}
\label{sec:exp:recon}
Here we evaluate the reconstruction performance of our inversion method, VARIN. First, we use \cref{alg:ar_inverse} to generate a set of inverse noises. These inverse noises are then used to reconstruct token maps at each scale. Finally, the final token maps are decoded back to images.

\begin{table}[h!]
    \centering
        \caption{Inversion reconstruction performance among discrete generative model. This indicates the underperformance of HART compared to Paelle in terms of reconstruction.}
    \label{tab:cv_reconstruction}
    \resizebox{1.0\linewidth}{!}{
    \setlength{\tabcolsep}{18pt}
    \begin{tabular}{llcccc}
    \toprule
    \multicolumn{2}{c}{Method}  & \multicolumn{4}{c}{Metric} \\
    \cmidrule(r){1-2} \cmidrule(l){3-6}
    Inversion &Model  & PSNR $\uparrow$     & LPIPS$_{^{\times 10^3}}$ $\downarrow$  & MSE$_{^{\times 10^4}}$ $\downarrow$     & SSIM$_{^{\times 10^2}}$ $\uparrow$         \\ 
    \midrule
    DICE &Paella  & 30.91 &39.81 &11.07 &90.22 \\
    VARIN &HART  & { 26.00} & {70.26} & {33.20} & {79.83}\\
    \bottomrule
    \end{tabular}
    }
\vspace{-0.2cm}
\end{table}

\noindent\textbf{Evaluation Metrics.}
\label{sec:exp:metric}
To measure the reconstruction, we adopt image similarity metric such as Peak Signal-to-Noise Ratio (PSNR), Learned Perceptual Image Patch Similarity (LPIPS) \citep{zhang2018unreasonable}, Mean Squared Error (MSE), and Structural Similarity Index Measure (SSIM) \citep{wang2004image} to compute the difference between source images and the reconstructed images.

\noindent\textbf{Results.}
In \cref{tab:cv_reconstruction}, we compares our inversion VARIN with DICE \cite{he2024dice}, as both operate on discrete distributions. While DICE is based on a discrete diffusion model, our approach utilizes HART, a discrete visual autoregressive model. Mathematically, discrete noise inversion for both DICE and VARIN should produce token maps identical to the input token maps. However, the recorded metrics reveal a gap between the reconstructed images and the original source images. This discrepancy arises because both Paella \citep{rampas2022novel} and HART rely on vector quantization autoencoders, which inherently involve reconstruction loss due to compression. From \cref{tab:cv_reconstruction}, we observe that the VAR-VAE of HART underperforms compared to the VQ-VAE used in Paella.

\subsection{Text-based Image Editing Performance} \label{sec:exp:editing}
This section demonstrates the effectiveness of our editing algorithm VARIN. For both Regeneration and VARIN, by default, we set the start scale of editing $s = 6$ based on observation from \cref{fig:hart} and set the $\tau$ to be $18$ in LAI. Furthermore, we propose a linear scheduler for $\lambda$ with respect to scale, where $\lambda=1$ on scale $s$ and start reducing linearly to $0$ at final scale $14$.

\begin{figure*}[t]
    \centering
    \includegraphics[width=1\linewidth]{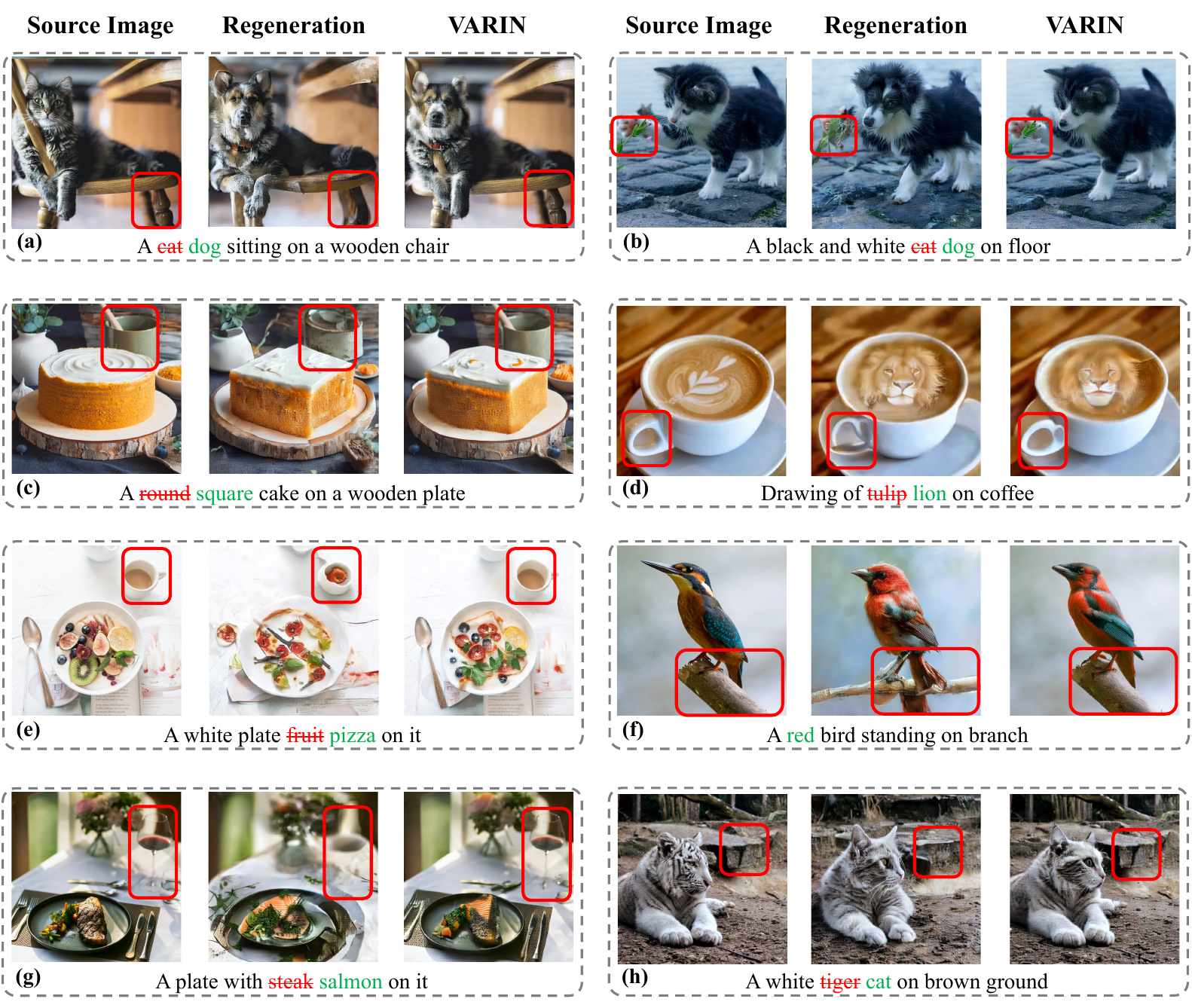}
    \vspace{-0.3cm}
    \caption{Qualitative result of editing results between VARIN and baseline Regeneration. We should better editing capability and background preservation.}
    \label{fig:main_qualitative}
    \vspace{-0.3cm}
\end{figure*}

\begin{figure*}[t]
    \centering
    \includegraphics[width=1\linewidth]{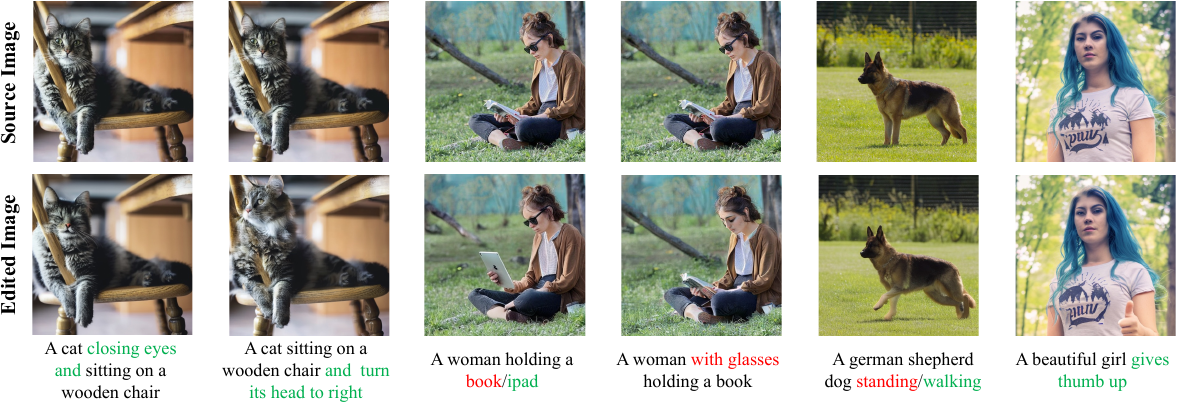}
    \vspace{-0.5cm}
    \caption{Fine-grained editing examples with VARIN using Switti \citep{voronov2024switti}. VARIN successfully performs localized modifications such as changing facial expressions, object replacement, and hand gestures while preserving original image details.}
    \label{fig:finegrain}
    \vspace{-0.5cm}
\end{figure*}

\noindent\textbf{Evaluation Metrics.} 
We evaluate the performance of our proposed editing methods on three main aspects: structural similarity, background preservation, and alignment between the edit prompt and the generated image.
To assess structural similarity between the original and generated images, we apply the structure distance metric from \cite{tumanyan2023plug}. For evaluating background preservation outside the editing mask, we use PSNR, LPIPS, MSE, and SSIM. Consistency between the edit prompt and the generated image is measured using the CLIP Similarity Score \citep{radford2021learning}, computed for the entire image and specifically for the regions within the editing mask.
These metrics offer a holistic assessment of our inversion method, addressing methods ability to maintain structure, preserve background details, and ensure prompt-image consistency, as in \cite{ju2023direct}.

\noindent\textbf{Results.}
Looking at \cref{tab:quantitative}, we begin by fairly comparing our editing method, VARIN, with training- and optimization-free approaches including Regeneration, DICE, and DDPM-Inversion, as VARIN similarly relies solely on an instant inversion technique. While VARIN outperforms DICE in editing effectiveness, it exhibits marginally lower preservation performance, likely due to HART’s weaker reconstruction capability compared to Paella (see \cref{sec:exp:recon}). Methodologically, VARIN is based on a next-scale autoregressive framework, whereas DICE adopts a discrete diffusion approach. Since the editing quality in both methods is closely tied to the sampling process, it is worth noting that VARIN requires only 10 sampling steps, while diffusion-based methods like DICE demand more, resulting in slower inference.
Empirically, VARIN achieves real-time editing at approximately 1 second per image, compared to ~2 seconds for DICE. Against DDPM-Inversion and Regeneration (see \cref{fig:main_qualitative} and \cref{fig:archs} in Supplementary Materials for qualitative comparisons with Regeneration), VARIN provides better reconstruction fidelity and background retention, while also adhering to the target prompt more effectively. In terms of efficiency, VARIN is approximately $10\times$ faster than DDPM-Inversion. Overall, among purely training-free editing methods, VARIN achieves competitive quantitative performance while substantially outperforming other diffusion based-approaches in editing speed. Secondly, within the category of autoregressive editing, we compare VARIN to the recently proposed EditAR model \citep{mu2025editar}. VARIN achieves superior performance across most evaluation metrics, with the exception of CLIP Edited Similarity. Unlike EditAR, which requires additional training akin to InstructPix2Pix \citep{brooks2023instructpix2pix}, VARIN relies solely on an inversion-based approach using next-scale autoregressive prediction, and does not require any task-specific fine-tuning.

In comparison to continuous diffusion-based editing techniques such as MasaCtrl \citep{cao_2023_masactrl}, MGIE \citep{fu2023guiding}, InstructPix2Pix \citep{brooks2023instructpix2pix}, Prompt-to-Prompt \citep{hertz2022prompt}, and Pix2Pix-Zero \citep{ramesh2021zero}, VARIN demonstrates superior background and structural preservation. It achieves a better balance between editing precision and content fidelity. This balance is also comparable to that of more advanced editing methods including InfEdit, PnP Inversion, Null-text Inversion, and Negative Prompting \citep{miyake2023negative}. Unlike these diffusion-based methods, VARIN enables real-time editing with an average inference time of approximately 1 second per image, leveraging instant inversion and next-scale autoregressive. In contrast, approaches such as Negative Prompting and Null-text Inversion often require time-consuming optimization during both inversion and editing stages. Other techniques like PnP and InfEdit depend on pretrained editing models (e.g., P2P) or attention manipulation, whereas VARIN is entirely training-free. While attention control and pretrained editing models offer promising directions, they are orthogonal to our approach and can be integrated with VARIN. We leave such extensions for future work. 

As demonstrated in \cref{fig:diverse}, VARIN effectively handles diverse prompts to add objects, change backgrounds, and alter image styles. These edits maintain alignment with the target prompts while preserving essential background details. Additionally, fine-grained editing results using the base model Switti \citep{voronov2024switti} are presented in \cref{fig:finegrain}, illustrating moderate local edits like closing eyes, turning heads, changing objects (e.g., book to iPad), and adjusting hand gestures or leg positions, though substantial pose or structural modifications remain challenging. In addition to the automated evaluations, we conducted a user study to assess and compare the visualization quality and editing prompt agreement of DDPM-Inv, DICE, and VARIN methods (detailed in \cref{tab:user_study} within the supplementary materials).
\vspace{-3mm}
\section{Conclusion}
\vspace{-2mm}
We proposed VARIN, that enable text-guided image editing capabilities for the text-to-image VAR (HART). Through extensive experiments, we have shown that this method successfully perform text-guided image editing while maintaining background preservation. These editing advancements extend HART's capabilities beyond mere text-to-image generation, making it a more versatile tool for real-world applications. Moving forward, promising directions for future research include exploring the application of VARIN to other traditional next-token autoregressive models or investigating attention control like \cite{hertz2022prompt} to further improve editing quality.

\newpage
\bibliography{main}
\bibliographystyle{abbrv}

\newpage
\appendix
\onecolumn
\section{Appendix}

In this supplementary material, we first present an ablation study of our proposed methods, Regeneration and VARIN. Subsequently, we discuss a variation of VARIN editing, referred to as only target editing.

\section{Regeneration Algorithm}
\begin{algorithm}[ht]
    \caption{Editing by Regeneration}
    \label{alg:regen}
    \begin{algorithmic}[1]
        \item[\textbf{Input:} Source image $I_\src$, text prompt $c_{\tgt}$]
        \State $r_1, r_2, \dots, r_K \leftarrow \mathcal{E}_\text{VAR}(I_{\src})$
        \For {$t$ from $s$ \text{to} $K$} \Comment{\textcolor{commentcolor}{$s$ is starting regeneration scale}}
        \State $\tilde r_t\sim p_\theta( \cdot |r_{<t}, c_{\tgt})$ \Comment{\textcolor{commentcolor}{Sampling using target prompt and token maps of previous scale}}
        \EndFor
        \State $I_{\tgt} \leftarrow \mathcal{D}_\text{VAR}(r_1, \dots, r_{s-}, \tilde r_s, \ldots, \tilde r_K)$
        \item[\textbf{Output:} $I_{\tgt}$]
    \end{algorithmic}
\end{algorithm}

\section{Ablation}
\label{sec:ablation}
In this section, we provide ablation details for our method VARIN and Regeneration. First, we examine the ablation study on the initial step of editing for the baseline Regeneration method. Finally, we do ablation on VARIN hyperparameter $\tau$ to indicate the importance of this hyperparameter in controlling how much information retains from the source images.

\subsection{Regeneration}
\begin{figure*}[ht]
    \centering
    \includegraphics[width=1\linewidth]{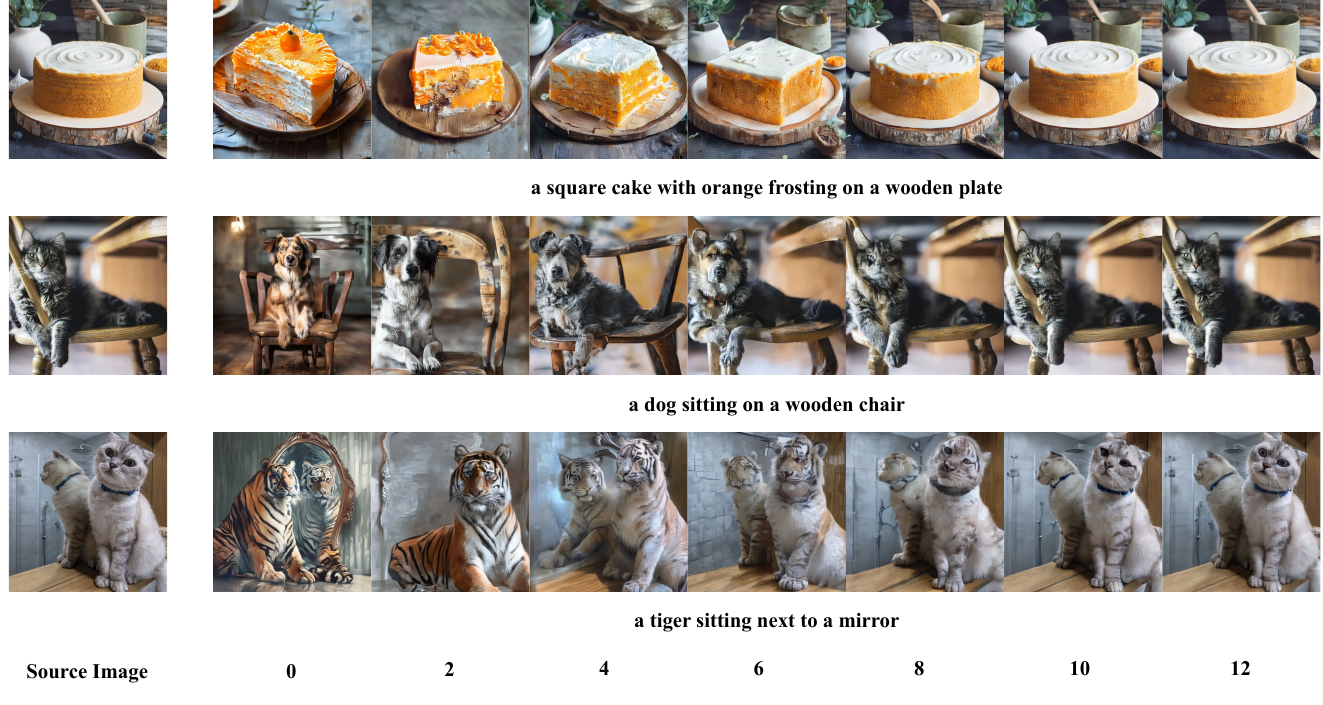}
    \caption{Ablation of Regeneration on beginning scale for editing $s$. We can see that as $s$ increases, the edited image are more like the source image. For small $s$, the edited image is too different from the source image. Therefore, the best $s$ for editing is around 6 to 8}
    \label{fig:regeneration_ablate}
\end{figure*}

For the Regeneration method, we preserve the first few scales (from $s = 0$ to $s=6$ or $s=7$) and generate the remaining scales using the target prompt. As shown in \cref{tab:ablate_regeneration}, increasing the starting editing scale $s$ leads to poorer performance on editing alignment metrics, such as structure distance and CLIP similarity, while improving performance on background preservation metrics. For qualitative results, refer to \cref{fig:regeneration_ablate}. When $s=0$, Regeneration behaves as a standard text-to-image generation process. However, when $s \geq 9$, the output image closely resembles the source image and, in some cases, fails to align with the provided target prompt (as illustrated in the first and second rows of \cref{fig:regeneration_ablate}). On the other hand, with smaller scales $s \leq 5$, preserving the background from the original image becomes significantly challenging. Therefore, for editing tasks, the most effective scale to begin editing is within the range of $s=6$ to $s=8$.

\begin{table*}[ht]
    \centering
    \resizebox{1.0\linewidth}{!}{
    \begin{tabular}{lccccccc}
    \toprule
    \multicolumn{1}{c}{Regeneration} &\multicolumn{1}{c}{Structure} &\multicolumn{2}{c}{CLIP Similarity} &\multicolumn{4}{c}{Background Preservation}\\
    \cmidrule(r){1-1} \cmidrule(lr){2-2} \cmidrule(l){3-4} \cmidrule(l){5-8}
      Begin Step& Distance$_{\times 10^3} \downarrow$ & Whole $\uparrow$ & Edited $\uparrow$ & PSNR $\uparrow$     & LPIPS$_{^{\times 10^3}}$ $\downarrow$  & MSE$_{^{\times 10^4}}$ $\downarrow$     & SSIM$_{^{\times 10^2}}$ $\uparrow$  \\
    \midrule
    \text{s = 6} &42.85 &26.83 &23.11 &19.21 &166.03 & 171.18 & {68.93}\\
    \text{s = 7} &33.70 &26.08 &22.57 &20.49 &137.00 &133.17 &70.79\\
    \text{s = 8} &25.56 &24.65 &21.13 &20.45 &106.50 &95.90 &73.31\\
    \bottomrule
    \end{tabular}}
    \caption{Ablation on beginning step to edit for method Regeneration. Similar to observation from \cref{fig:regeneration_ablate}, as $s$ increases, the target alignment becomes worse while the background preservation is better. The recommended scale $s$ is 6, based on qualitative result.}
    \label{tab:ablate_regeneration}
\end{table*}

\subsection{VARIN}

\begin{figure*}[ht]
    \centering
    \includegraphics[width=\linewidth]{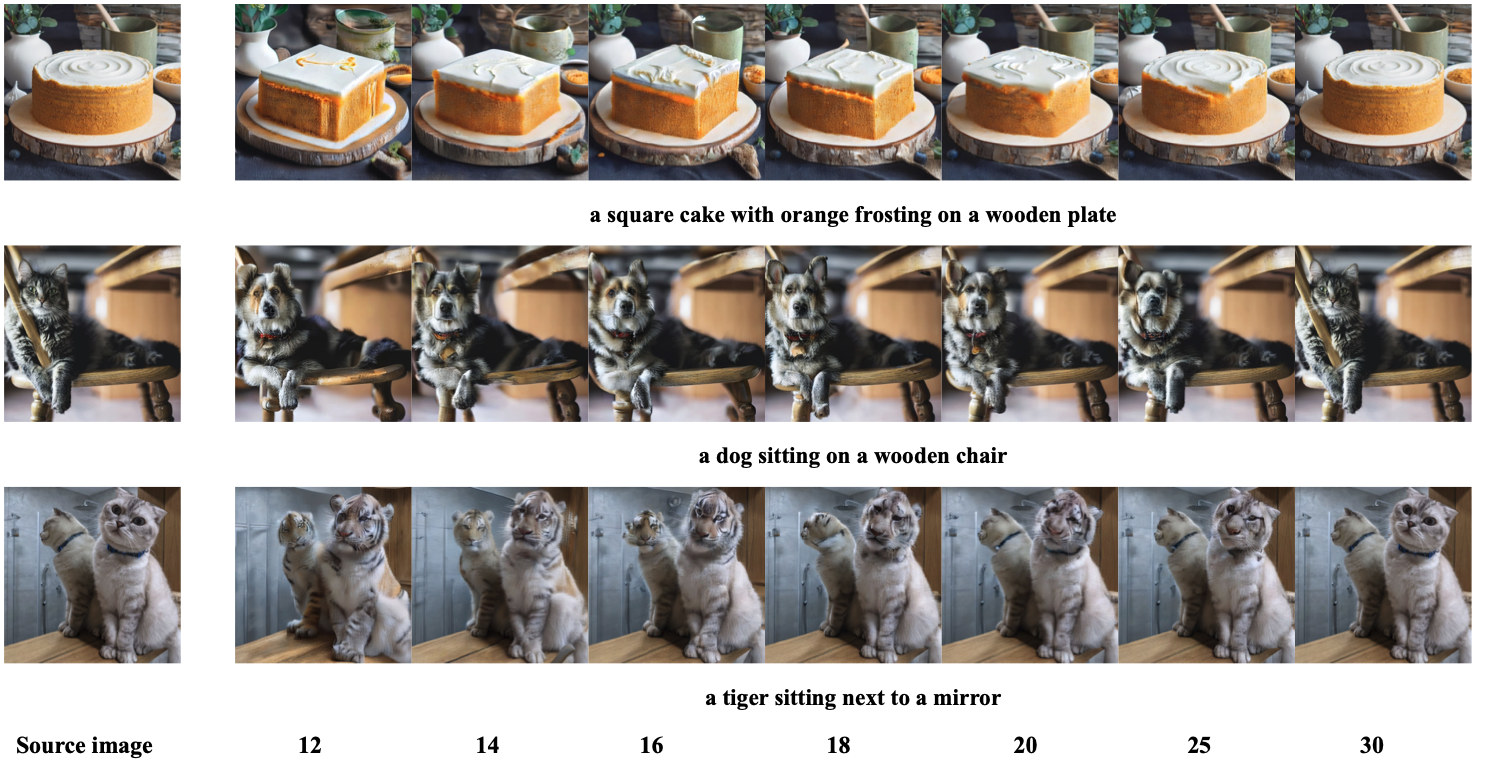}
    \caption{Qualitative result for ablation of $\tau$ for VARIN.}
    \label{fig:tau}
\end{figure*}

Observing the qualitative and quantitative results of the Regeneration technique, we select the beginning editing scale $s = 6$. As mentioned in main paper, $\tau$ in \cref{algo:LAI} also controls the retention of source information in the inverse noise $n_t$. As $\tau$ increases, the output image preserves the background more effectively (refer to \cref{fig:tau}). Without $\tau$, controlling the editing process becomes challenging, as inverse noise contains greater uncertainty about the source information, making it more sensitive and less suitable for editing controllability. Since the hyperparameter $\tau$ is crucial for editing, we perform an ablation study on it in \cref{tab:linear_source}. Our findings indicate that $\tau$ values between $14$ and $20$ produce the best visual editing results.

\begin{table*}[ht!]
    \centering
    \resizebox{1.0\linewidth}{!}{
    \begin{tabular}{lccccccc}
    \toprule
    \multicolumn{1}{c}{$\text{VARIN}$} &\multicolumn{1}{c}{Structure} &\multicolumn{2}{c}{CLIP Similarity} &\multicolumn{4}{c}{Background Preservation}\\
    \cmidrule(r){1-1} \cmidrule(lr){2-2} \cmidrule(l){3-4} \cmidrule(l){5-8}
      $\tau$ & Distance$_{\times 10^3} \downarrow$ & Whole $\uparrow$ & Edited $\uparrow$ & PSNR $\uparrow$     & LPIPS$_{^{\times 10^3}}$ $\downarrow$  & MSE$_{^{\times 10^4}}$ $\downarrow$     & SSIM$_{^{\times 10^2}}$ $\uparrow$  \\
    \midrule
    14 &19.21	&25.73	&22.12	&23.97	&74.23	&68.49	&82.26\\
    16 &14.33	&25.32	&21.72	&25.49	&60.87	&48.97	&84.34\\
    18 &11.46	&25.05	&21.49	&26.54	&54.04	&38.33	&85.39\\
    20 &9.85	&24.79	&21.28	&27.30	&50.28	&31.78	&85.96\\
    \bottomrule
    \end{tabular}}
    \caption{Ablation on value $\tau$.}
    \label{tab:linear_source}
\end{table*}

\section{Gumbel Truncation Sampling}
For the Gumbel truncation algorithm \cite{kool2019stochastic}, we provide the detailed algorithm in~\cref{alg:gumbel_trunc}.

\begin{algorithm}[h!]
    \caption{GumbelTrunc}
    \label{alg:gumbel_trunc}
    \begin{algorithmic}[1]
        \item[\textbf{Input:} location $\phi$, threshold $T$]
        \State $u \sim \mathrm{Uniform}(0, I)$
        \item[\textbf{Return:} $\phi - \log(\exp(\phi - T) - \log u)$]
    \end{algorithmic}
\end{algorithm}

\section{Editing using Only Target Prompt}

\begin{figure}[h!]
    \centering
    \includegraphics[width=0.5\linewidth]{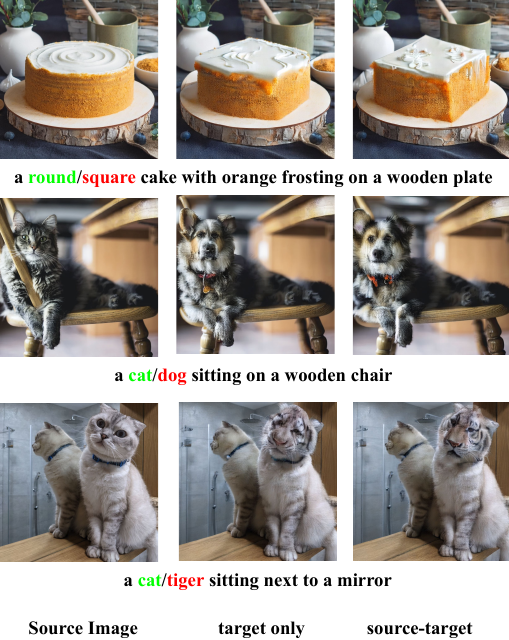}
    \caption{The first column is source image. The second column is target VARIN \cref{alg:src_edit}, and the third column is source-target VARIN \cref{alg:edit_noise}. The green and red color texts are source and target prompt, correspondingly.}
    \label{fig:2edit}
\end{figure}

In this section, we discuss the only target prompt VARIN, a variant of source-target VARIN in the main paper. In this editing algorithm, we only use target prompt $c_{\tgt}$ for both noise extraction and editing. First, we extract a set of inverse noises using the target prompt. This extracted noise set can then be used to perform the editing process, as shown in \cref{alg:src_edit}. For this algorithm, qualitative results are provided in \cref{fig:2edit}, demonstrating that the source-target VARIN performs well on the editing task. It is worth noting that for only target VARIN, the effective $\tau$ value is lower, typically between $10$ and $14$. In \cref{tab:linear}, we provide the ablation on $\tau$ for only target VARIN editing method.

\begin{algorithm}[h!]
    \caption{Editing by target VARIN}
    \label{alg:src_edit}
    \begin{algorithmic}[1]
        \item[\textbf{Regeneration:}]
        \State $r_1, r_2, \dots, r_K \leftarrow \mathcal{E}_\text{VAR}(I_{\tgt})$
        \State $n_1, n_2, \dots, n_K \leftarrow \text{VARIN}((r_1, r_2, \dots, r_K), c_{\tgt})$
        \For {$t$ from $s$ \text{to} $K$} \Comment{\textcolor{commentcolor}{$s$ is the scale we start editing}}
        \State $p_t\leftarrow p_\theta(r_t|r_{<t}, c_{\tgt})$ 
        \Comment{\textcolor{commentcolor}{$p_t$ is log probability}}
        \State $g \sim \text{Gumbel}(0, I)$
        \State $q_t = p_t + (1-\lambda)\cdot g + \lambda\cdot n_t $
        \State $r_t = \text{argmax}(q_t)$
        \EndFor
        \State $I_{tgt} \leftarrow \mathcal{D}_\text{VAR}(r_1, r_2, \dots, r_K)$
        \State Return $I_{tgt}$.
    \end{algorithmic}
\end{algorithm}

\begin{table*}[ht]
    \centering
    \resizebox{1.0\linewidth}{!}{
    \begin{tabular}{lccccccc}
    \toprule
    \multicolumn{1}{c}{VARIN} &\multicolumn{1}{c}{Structure} &\multicolumn{2}{c}{CLIP Similarity} &\multicolumn{4}{c}{Background Preservation}\\
    \cmidrule(r){1-1} \cmidrule(lr){2-2} \cmidrule(l){3-4} \cmidrule(l){5-8}
      $\tau$ & Distance$_{\times 10^3} \downarrow$ & Whole $\uparrow$ & Edited $\uparrow$ & PSNR $\uparrow$     & LPIPS$_{^{\times 10^3}}$ $\downarrow$  & MSE$_{^{\times 10^4}}$ $\downarrow$     & SSIM$_{^{\times 10^2}}$ $\uparrow$  \\
    \midrule
    8 &15.00	&25.54	&21.96	&25.34	&62.62	&50.81	&83.96\\
    10 &11.76	&25.17	&21.63	&26.59	&54.45	&38.04	&85.30\\
    12 &9.91	&24.93	&21.37	&27.41	&50.38	&31.35	&85.93\\
    14 &8.84	&24.74	&21.23	&27.86	&48.19	&28.32	&86.26\\
    \bottomrule
    \end{tabular}}
    \caption{Ablation on value $\tau$ of $\text{VARIN}$. As the $\tau$ increases, the edited image is more like the source image and preserve background better}
    \label{tab:linear}
\end{table*}

\begin{figure}[h]
    \centering
    \includegraphics[width=1.0\linewidth]{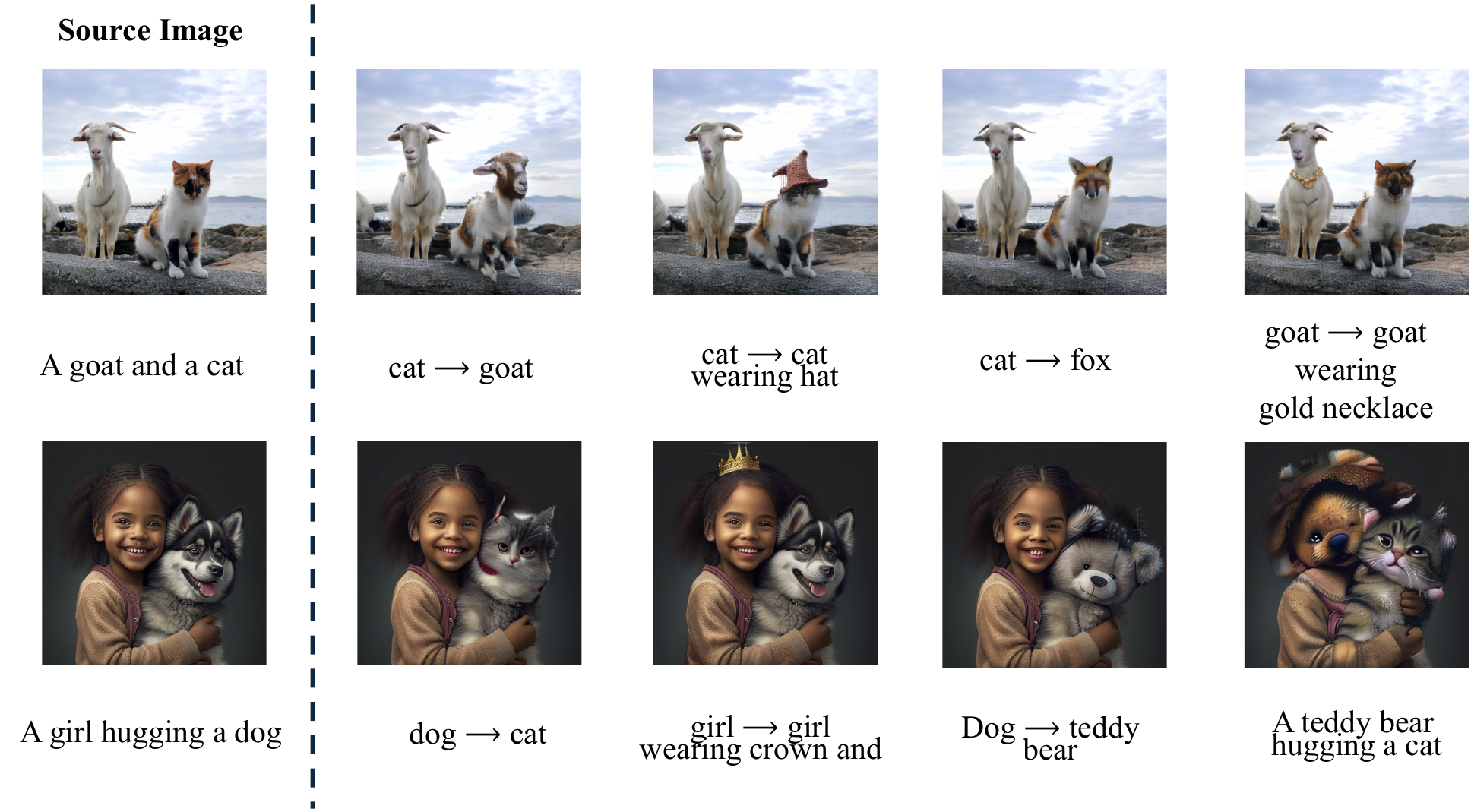}
    \caption{Editing results on complex scene involves two objects.}
    \label{fig:2obj}
\end{figure}


\begin{figure}[h]
    \centering
    \includegraphics[width=1.0\linewidth]{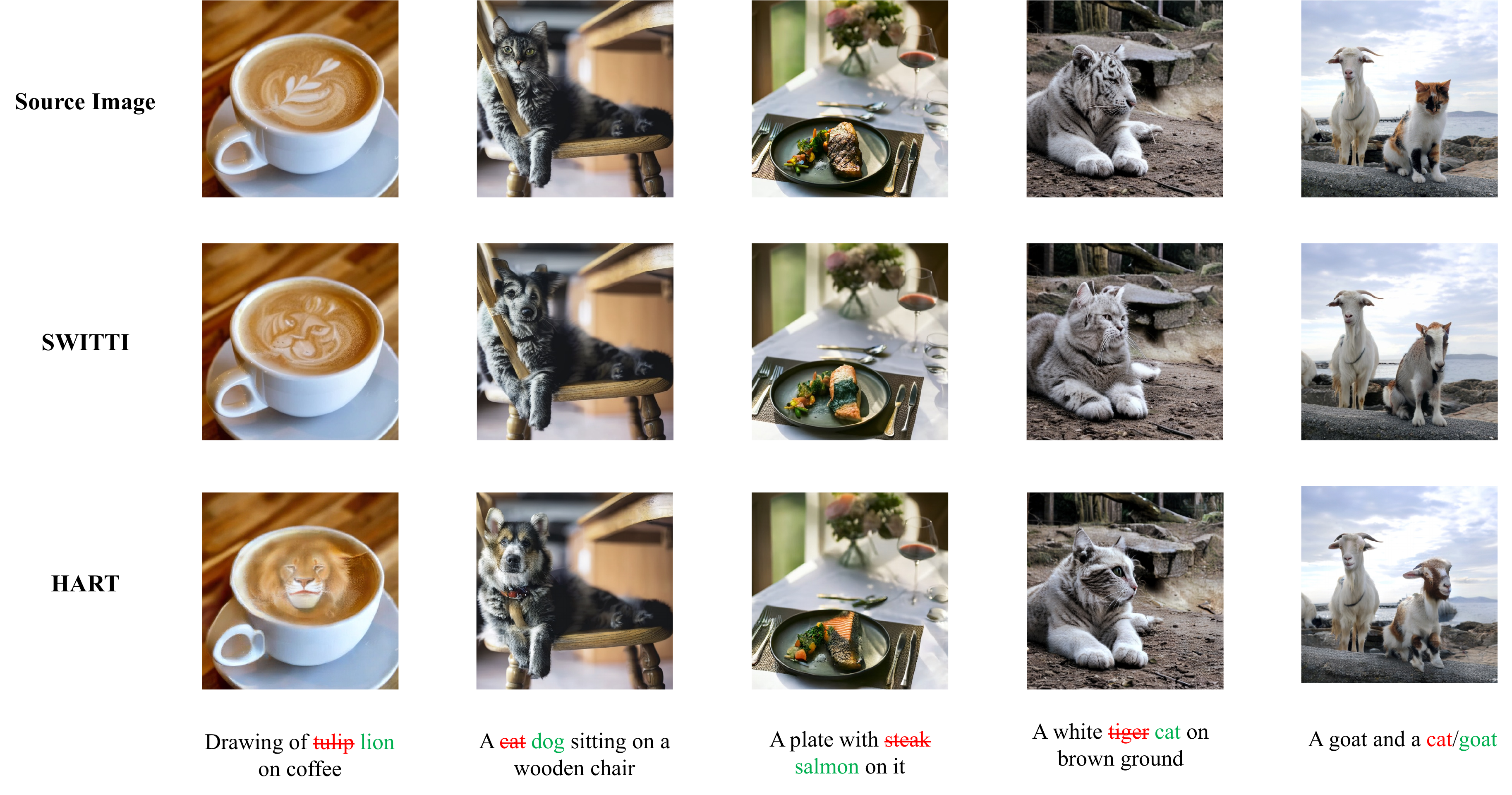}
    \caption{Extending VARIN to different architectures and base models.}
    \label{fig:archs}
\end{figure}

\begin{figure}[h]
    \centering
    \includegraphics[width=1.0\linewidth]{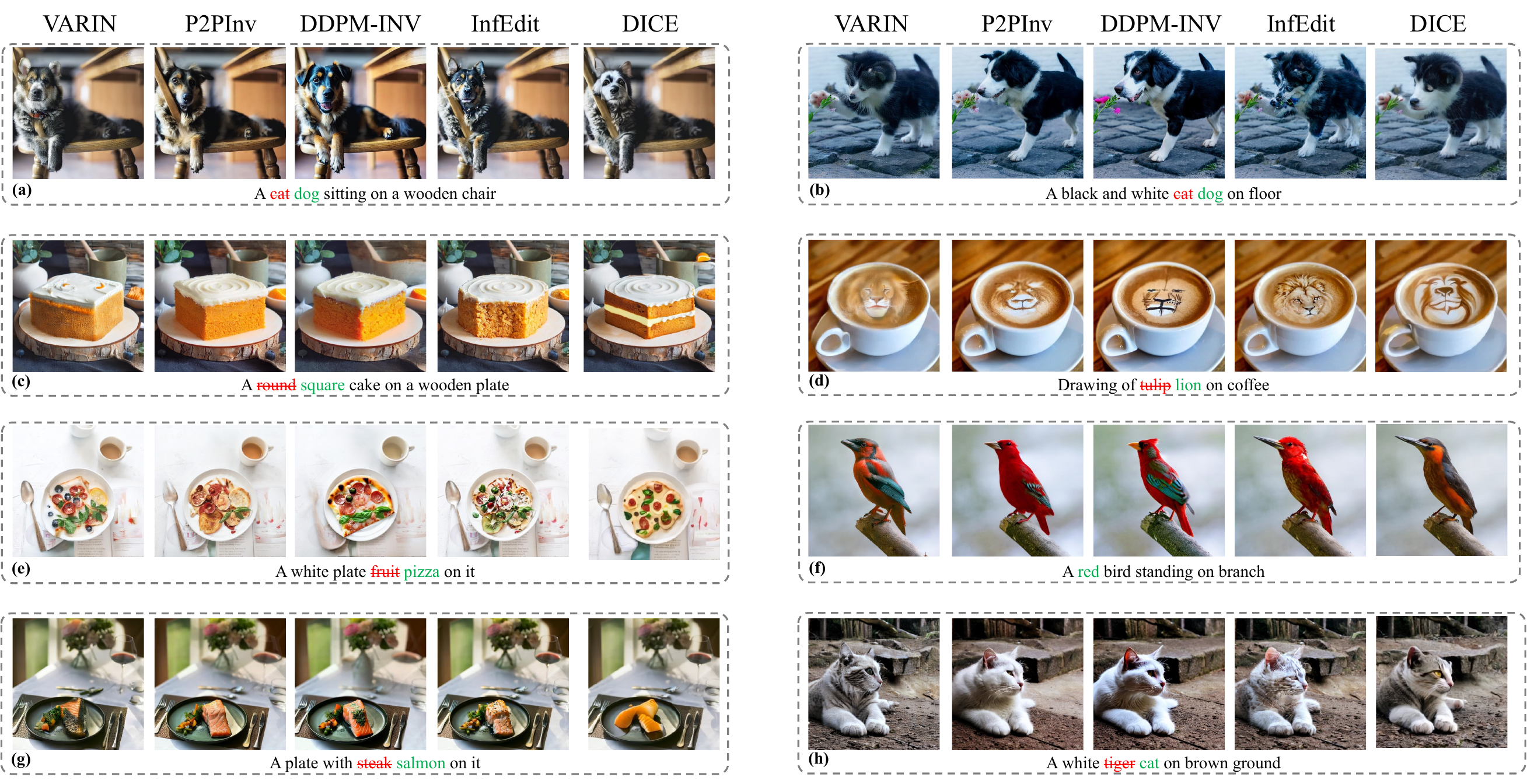}
    \caption{Comparing VARIN with different methods.}
    \label{fig:compare}
\end{figure}

\begin{table}[h]
    \centering
    \begin{tabular}{lccc}
    \toprule
        & DDPM-Inv & DICE & VARIN\\
    \midrule
    Visualization quality& 35.89\% & 15.89\% & 48.24\% \\
    \midrule
    Agreement to the editing prompt & 52.35\% & 41.76\% & 64.71\%\\
    \bottomrule \\
    
    \end{tabular}
    \caption{User Study conducting on 25 people and 10 images.}
    \label{tab:user_study}
\end{table}

\begin{figure}[h]
    \centering
    \includegraphics[width=0.6\linewidth]{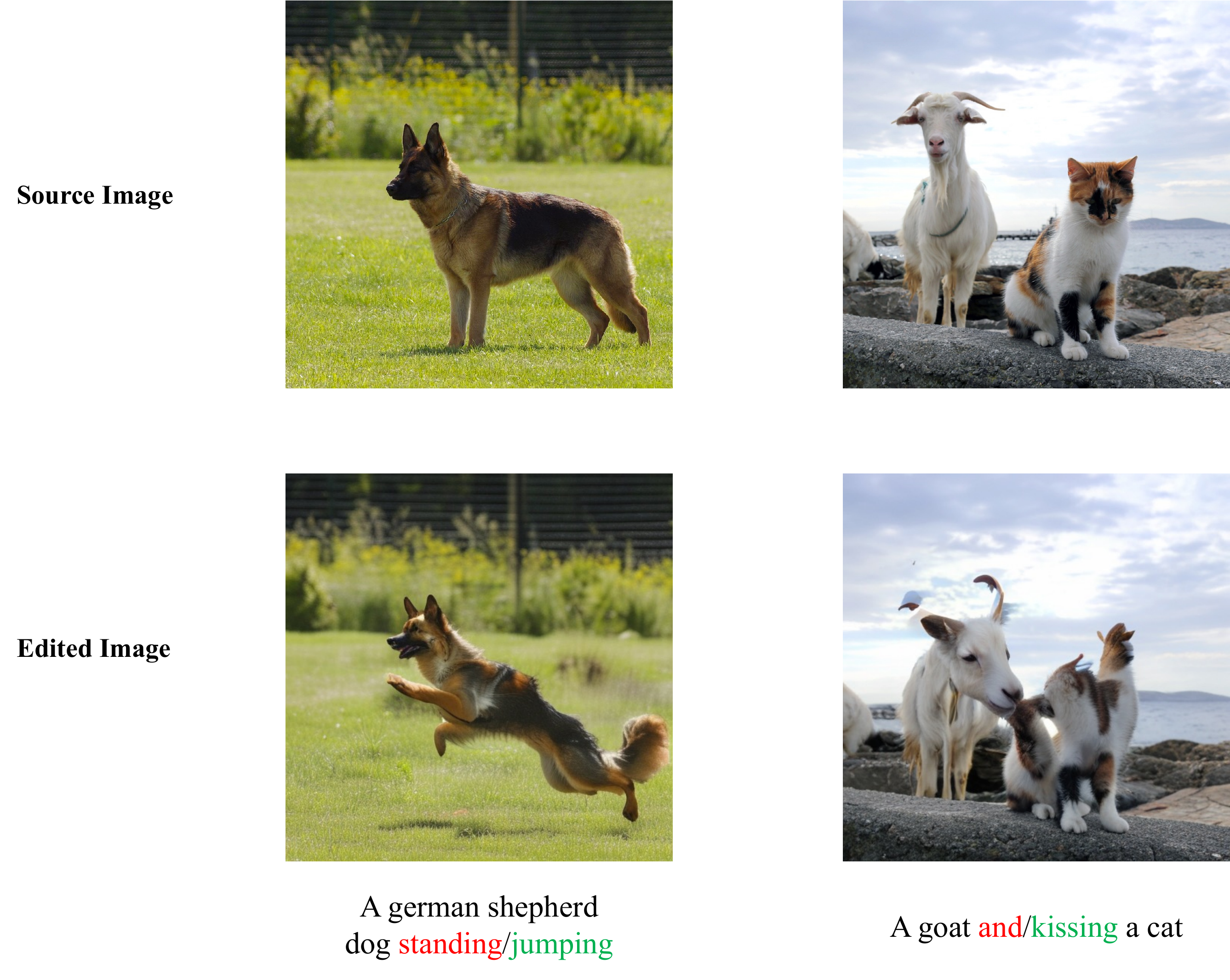}
    \caption{Failure cases: large movement and complex interaction between object}
    \label{fig:compare}
\end{figure}

\section{More Qualitative Comparison}
See \cref{fig:2obj},~\cref{fig:archs}, and \cref{fig:compare}.

\section{Limitation}
Here we demonstrate one of the failure case that serves the limitation of our proposed method. As demonstrated in \cref{fig:compare},  our method may fail in cases involving large pose or structural changes, or complex interactions that require such changes. These scenarios can challenge the model’s ability to preserve consistency and produce realistic edits. 
\clearpage

\end{document}